\documentclass[journal,10pt]{IEEEtran}
\usepackage{cite}
\usepackage{amsmath,amssymb,amsfonts}
\usepackage{algorithmic}
\usepackage{algorithm}
\usepackage{graphicx}
\usepackage{textcomp}
\usepackage{xcolor}
\usepackage{balance}
\usepackage{url}
\usepackage{hyperref}

\usepackage[labelsep=colon, singlelinecheck=false]{caption}
\begin{document}

% Example definitions.
% --------------------
\def\b{{\mathbf b}}
\def\c{{\mathbf c}}
\def\v{{\mathbf v}}
\def\x{{\mathbf x}}
\def\y{{\mathbf y}}
\def\X{{\mathbf X}}
\def\Y{{\mathbf Y}}
\def\F{{\mathbf F}}
\def\G{{\mathbf G}}
\def\H{{\mathbf H}}
\def\W{{\mathbf W}}
\def\A{{\mathbf A}}
\def\B{{\mathbf B}}
\def\C{{\mathbf C}}
\def\M{{\mathbf M}}
\def\Z{{\mathbf Z}}
\def\U{{\mathbf U}}
\def\V{{\mathbf V}}
\def\I{{\mathbf I}}
\def\w{{\mathbf w}}
\def\Q{{\mathbf Q}}
\def\S{{\mathbf \Sigma}}
\def\SigmaW{{\mathbf \sigma_\v}}
\def\L{{\cal L}}

\newcommand{\temp}[1]{\textcolor{red}{#1}}
\newcommand{\mario}[1]{\textcolor{green}{#1}}

\title{Latent Space Alignment for AI-Native \\ MIMO Semantic Communications\vspace{.2cm}}
% 1\textsuperscript{st} 
\author{Mario Edoardo Pandolfo$^{1,2}$, Simone Fiorellino$^{1,2}$, Emilio Calvanese Strinati$^3$, and Paolo Di Lorenzo$^{2,4}$ \medskip \\
$^1$ DIAG Department, Sapienza University of Rome, via Ariosto 25, Rome, Italy.\\
$^2$ Consorzio Nazionale Interuniversitario per le Telecomunicazioni (CNIT), Parma, Italy.\\
$^3$ CEA Leti, University Grenoble Alpes, 38000, Grenoble, France.\\
$^4$ DIET Department, Sapienza University of Rome, Via Eudossiana 18, Rome, Italy. \smallskip\\
E-mail: \{marioedoardo.pandolfo, simone.fiorellino, paolo.dilorenzo\}@uniroma1.it, emilio.calvanese-strinati@cea.fr.
\thanks{\hrule\vspace{.15cm}
This work was funded by the 6G-GOALS project under the 6G SNS-JU Horizon program, n.101139232, and by the European Union under the Italian National Recovery and Resilience Plan of NextGenerationEU, partnership on “Telecommunications of the Future” (PE00000001 - program “RESTART”).}\vspace{-.5cm}}

%\author{\IEEEauthorblockN{Anonymous Authors}} % TODO

\maketitle

\begin{abstract}
Semantic communications focus on prioritizing the understanding of the meaning behind transmitted data and ensuring the successful completion of tasks that motivate the exchange of information. However, when devices rely on different languages, logic, or internal representations, semantic mismatches may occur, potentially hindering mutual understanding. This paper introduces a novel approach to addressing latent space misalignment in semantic communications, exploiting multiple-input multiple-output (MIMO) communications. Specifically, our method learns a MIMO precoder/decoder pair that jointly performs latent space compression and semantic channel equalization, mitigating both semantic mismatches and physical channel impairments. We explore two solutions: (i) a linear model, optimized by solving a biconvex optimization problem via the alternating direction method of multipliers (ADMM); (ii) a neural network-based model, which learns semantic MIMO precoder/decoder under transmission power budget and complexity constraints. Numerical results demonstrate the effectiveness of the proposed approach in a goal-oriented semantic communication scenario, illustrating the main trade-offs between accuracy, communication burden, and complexity of the solutions.
\end{abstract}
\smallskip

\begin{figure*}[t]
    \centering
    \includegraphics[width=0.98\textwidth, trim=2bp 1bp 15bp 0bp, clip]{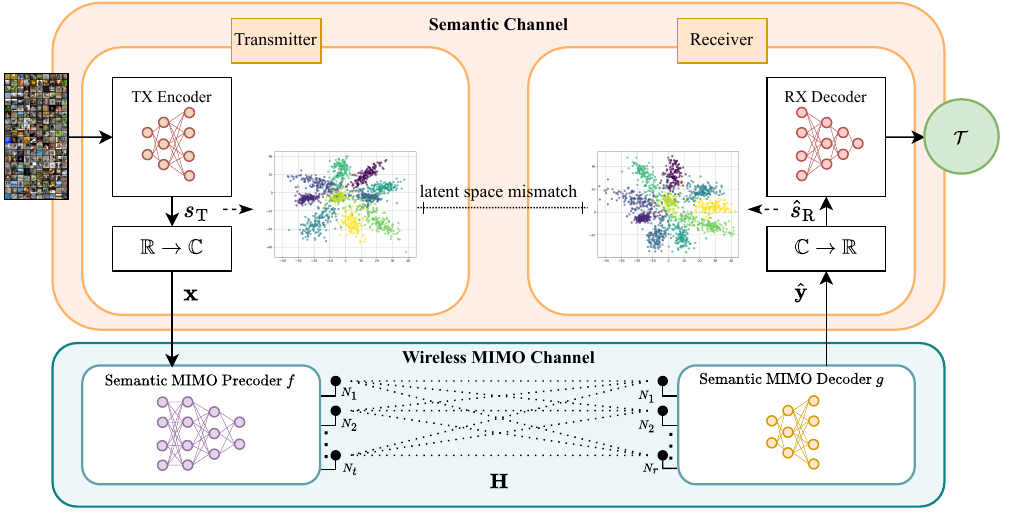}
    %\vspace{-0.1cm}
    \caption{System model: The semantic precoder $f$ of TX encodes a compressed latent space $\mathbf{s_i}$ into a transmitter vector of dimension $N_T$. It is transmitted through the MIMO channel $\H$ with noise $\v$. Then, the received vector is decoded by the semantic decoder $g$ and decompressed to match the RX latent space.}
    \label{fig:system_model}
\end{figure*}

\vspace{-.4cm}
\section{Introduction}
\label{sec:intro}

Classic communication paradigms focus on the accurate transmission of individual symbols or bits. The rapidly increasing number of connected devices, combined with the sheer volume of information required by emerging applications, is rapidly overwhelming the capacity of traditional bit-centric communication systems \cite{alwis2021devices, shi2021new, li2018Beyond}. For instance, in time-sensitive applications like autonomous driving, industrial automation, and smart surveillance, it is crucial to deliver reliable AI services with ultra-low latency while balancing energy efficiency, communication overhead, and computational capacity. 
Despite ongoing advancements, current wireless systems are increasing bandwidth and energy use to meet the growing demand for higher data rates. However, this pursuit faces resource constraints, including limited spectrum, energy, and computational power, highlighting the need for a new communication paradigm for future applications \cite{strinati20216g,getu2023semantic,strinati2024goal,gunduz2022beyond,luo2022semantic}.

A key paradigm shift for 6G systems and envisioned IoT applications lies in the adoption of semantic communications (SCs) \cite{strinati20216g,strinati2024goal}. This approach moves beyond traditional bit-related metrics commonly used in current system design and optimization, instead prioritizing the interpretation of the meaning behind transmitted data and the successful completion of tasks that drive the information exchange  \cite{guler2018semantic, bao2011towards, gunduz2022beyond,strinati2024goal}. % SOTA in Semantic Com
Taking this idea further, novel approaches to semantic joint source and channel coding \cite{gunduz2022beyond}, semantic extraction and compression \cite{kountouris2021semantics, stavrou2023role}, goal-oriented system design and optimization \cite{binucci2022adaptive,di2023goal,fiorellino2024dynamic}, semantic reasoning \cite{thomas2023reasoning}, and semantic communications based on generative AI \cite{barbarossa2023semantic} have been proposed in the literature. A key enabler of SCs is artificial intelligence (AI), where deep neural networks (DNNs) are exploited to extract and transmit key semantic features from data rather than the raw data itself \cite{xie2021deep,gunduz2022beyond}. At the receiver side, a DNN decoder reconstructs and interprets the intended message, offering inherent robustness to noise and imperfect transmission conditions. This approach was shown to reduce the amount of data that needs to be transmitted, thereby saving bandwidth and energy \cite{xie2021deep,gunduz2022beyond}.
% \\ % PROBLEM
In SCs, two primary sources of noise exist \cite{luo2022semantic}: (i) \textit{channel noise}, which alters the received symbols at the physical layer; (ii) \textit{semantic noise}, which can arise from various causes, e.g., misalignment of logic, interpretation, or knowledge among communication devices. Both types of noise can significantly disrupt communication.

The handling of channel noise has historically been guided by Shannon's Separation Theorem \cite{shannon1948mathematical}, which asserts that separate source and channel coding achieves optimality in the asymptotic regime of infinite block length and unbounded complexity. Although suitable for traditional communication networks, this approach has proven suboptimal in practical finite block-length scenarios, especially for emerging applications. Joint source-channel coding (JSCC) addresses these limitations by directly mapping source signals to channel symbols, bypassing intermediate bit representations. Recent advances in deep learning have led to practical deep learning-based JSCC (DeepJSCC) schemes \cite{8723589, 9714510, 9438648, 9791398, 10597355}, where DNNs are used to jointly optimize the source-channel coding problem in an end-to-end fashion.
DeepJSCC shown to be suitable for enabling SCs schemes \cite{gunduz2022beyond, 10328187} while performing both semantic compression and channel equalization (source-channel coding). However, most existing DeepJSCC frameworks assume consistent latent representations at transmitter and receiver, and therefore do not address the issue of semantic noise.
\\
\indent In this work, we specifically address the challenge of \textit{latent space misalignment}, i.e., a type of semantic noise due to the heterogeneous ways used by DNNs at different devices to encode and interpret the exchanged semantic information. This problem, also known as \textit{semantic channel equalization}, was investigated in some recent works \cite{alvarez2019towards,moschella2022relative,pan2023stitchable,sana2023semantic,fiorellino2024dynamic,lahner2024direct,maiorca2024latent,huttebraucker2024relative}. In \cite{sana2023semantic}, the authors introduce a semantic channel equalizer that models mismatches with measurable transformations over semantic spaces and applies tools from optimal transport theory. The work in \cite{moschella2022relative} introduces relative representations (RRs) to perform (zero-shot) latent space communication among DNN models, and the paper in \cite{fiorellino2024dynamic} exploits RRs in a goal-oriented communication context performing a dynamic optimization of edge resources. 
In \cite{maiorca2024latent}, the authors leverage RRs to align models without the need to train a new decoder. The alignment process is achieved by computing the pseudo-inverse of the anchor matrix, and cosine similarity is used as the similarity function.
In \cite{huttebraucker2024relative} is proposed a semantic equalization framework based on RRs that do not need any retraining and require sharing a small amount of data between transmitter and receiver. The method is agnostic to the similarity function of RRs.
However, the works in \cite{sana2023semantic,fiorellino2024dynamic,moschella2022relative,maiorca2024latent,huttebraucker2024relative} did not investigate the effect and, most importantly, the optimization of MIMO wireless communication for semantic channel equalization, thus leading to a full integration of AI-based semantic extraction and wireless aspects.
\\
\indent \textbf{Contributions.} In this work, we present an optimization framework for AI-native, MIMO, semantic, and wireless communications, addressing the impact of semantic noise modeled as latent space mismatch.
Specifically, we model the MIMO communication as a chain of (learnable) transformations composed of a (semantic) precoder, the channel matrix, and a (semantic) decoder. Then, we perform a joint optimization of semantic precoding/decoding that minimizes a semantic performance metric of interest, promoting latent space alignment between semantic feature spaces at the transmitter and receiver sides. To trade-off complexity and performance, we explore two optimization solutions. The first assumes linear precoding/decoding modules, and the related problem is formulated as a biconvex program numerically solved through ADMM. In the second case, we assume nonlinear precoding/decoding modules modeled by DNNs, whose parameters are learned through a specific training procedure, which takes into account transmission power budget and complexity constraints. The proposed framework performs semantic compression and equalization jointly, while handling both semantic and channel noise effectively. Additionally, a detailed analysis is carried out to assess the computational complexity of the proposed DNN solution. Numerical results evaluate the effectiveness of our methodology in a goal-oriented semantic communication setting, where data exchange supports image classification, illustrating how our approach leads to better performance with respect to available benchmarks from the literature.

\section{System Model}
\label{sec:case_obj}

We consider a communication system where the transmitter (TX) and the receiver (RX) employ (pre-trained) DNNs to encode and decode semantic information, resulting in the transmission (and interpretation) of latent representations. A pictorial representation is illustrated in Fig. \ref{fig:system_model}. Let $\mathbf{s}_T\in \mathbb{R}^d$ be the vector of semantic features extracted at the TX side from a data point $\mathbf{z}\in \mathbb{R}^q$. The set of all semantic vectors $\mathbf{s}_{T}$ represents the TX semantic latent space. We assume that the RX is trained to interpret a different encoding scheme than the one used by the TX, i.e., it requires the reception of a different semantic feature vector $\mathbf{s}_{R}\in \mathbb{R}^m$ (corresponding to $\mathbf{z}$) to correctly interpret the transmitted message or effectively perform a given task (e.g., classification). The set of all semantic vectors $\mathbf{s}_{R}$ represents the RX semantic latent space. Due to the mismatches between TX and RX latent spaces, the communication is prone to semantic noise and must be equalized to enable understanding between TX and RX. 
\\
Our approach to semantic equalization exploits the presence of MIMO wireless channels. Thus, the first step is to build a complex semantic latent space that is more suitable for efficient MIMO communication. Assuming, without loss of generality, that $d$ is even, this approach involves pairing the first half of the semantic features in $\mathbf{s}_T\in \mathbb{R}^d$ with the second half to form complex symbols, yielding an input vector $\x\in\mathbb{C}^\frac{d}{2}$. Then, assuming the TX is endowed with $N_T$ antennas, we exploit a \textit{semantic MIMO precoder} that maps the complex vector $\x\in\mathbb{C}^\frac{d}{2}$ into the vector $\overline{\mathbf{x}}\in\mathbb{C}^{K N_T}$ to be transmitted over $K$ wireless channel usages. The semantic MIMO precoder implements a learnable transformation represented by the function $f: \mathbb{C}^\frac{d}{2}\to \mathbb{C}^{K N_T}$. This transformation additionally enables the semantic compression of the TX latent space (since we typically consider $\frac{d}{2} \gg K N_T$), with a compression factor defined as:
\begin{align}\label{eq:compression_factor}
\zeta = \frac{K N_T}{d/2}.
\end{align}
We assume a MIMO flat Rayleigh fading channel described by the matrix $\overline{\mathbf{H}}\in\mathbb{C}^{N_T\times N_R}$, where $N_R$ denotes the number of antennas at the RX side, and each element $h_{ij}$ in $\overline{\H}$ is modeled as a zero-mean complex Gaussian random variable representing the fading effect between the $i$-th transmit antenna and the $j$-th receive antenna. We assume the channel to be constant within the time span by $K$ consecutive transmissions. Then, at the RX side, we have a \textit{semantic MIMO decoder} that maps the received symbols into a complex vector $\y\in\mathbb{C}^\frac{m}{2}$, via the learnable transformation $g: \mathbb{C}^{K N_R}\to \mathbb{C}^{\frac{m}{2}}$. Overall, the considered semantic MIMO communication channel can be compactly written as:
\begin{align}\label{eq:problem_setting}
    \hat{\y} = g(\H f(\x) + \v),
\end{align}
where $\H =\I_K \otimes \overline{\H}\in\mathbb{C}^{KN_R\times KN_T}$, with $\I_K$ denoting the $K\times K$ identity matrix, and $\otimes$ being the Kronecker product; also, $\v\in\mathbb{C}^{K N_R}$ represents the receiver noise vector, which follows a distribution $\mathcal{CN}(\mathbf{0},\Sigma_\v)$. Finally, the complex latent vector $\hat{\y}\in \mathbb{C}^{\frac{m}{2}}$ in (\ref{eq:problem_setting}) is converted into an $m$-dimensional real latent vector, say $\hat{\mathbf{s}}_{\text{R}}$, by inverting the halving operation done at the TX side. Our aim is to act on the learnable transformations $f$ and $g$ in (\ref{eq:problem_setting}) to perform the best possible alignment between TX and RX semantic latent spaces. This can be obtained by minimizing the (semantic) distance between the spaces composed by the vectors $\mathbf{s}_{R}$ and $\hat{\mathbf{s}}_{R}$ over a training set $\mathcal{P}$ of available data points, which represent \textit{semantic pilots} (SPs) in our context. In the sequel, we will illustrate two methods to perform and optimize semantic channel equalization based on the model (\ref{eq:problem_setting}), considering different complexities and imposing constraints on the TX power.

%\footnote{This operation can also be implemented splitting the vector $\mathbf{x}$ into $p$ blocks of $N_T$ elements, and applying separable transformations to each block.}. 

\section{Joint Optimization of Semantic Precoder and Decoder for Latent Space Alignment}

In this section, we introduce two methods for joint optimization of semantic precoding/decoding to enable latent space alignment between semantic feature spaces at the TX and RX sides. The first one assumes linear precoding/decoding modules, whereas the second method exploits nonlinear precoding/decoding modules modeled by DNNs. In the sequel, we assume perfect estimation of the MIMO channel matrix $\H$; the effect of imperfect channel state information (CSI) will be investigated in our future works.
\vspace{-.4cm}

\subsection{Linear Semantic Precoding and Decoding}
\label{sec:linear_formulation}

Here, we focus on the linear case, which 
provides a simplified yet effective approach for modeling semantic encoding and decoding functions, allowing for tractable optimization. Specifically, the semantic encoding function, $f$, is modeled as a linear transformation, represented by matrix $\F\in\mathbb{C}^{K N_T\times \frac{d}{2}}$, and the decoding function, $g$, is modeled by matrix $\G\in\mathbb{C}^{\frac{m}{2}\times K N_R}$. Here, w.l.o.g., we consider the application of \textit{pre-whitening} to standardize the covariance of the transmitted symbols ($\x$) to the identity matrix. This assumption helps formulate a constraint on the TX power that is expressed solely in terms of $\F$, i.e., $\text{tr}(\F\F^H)\le P_T$, with $P_T>0$ denoting the maximum power budget. Hence, \eqref{eq:problem_setting} can be cast as:
\begin{equation}\label{eq:linear_MIMO_x}
  \hat{\y}=\G\H\F\x +\G\v.
\end{equation}
Our optimization criterion aims at minimizing the (semantic) distance between the signals given by model \eqref{eq:linear_MIMO_x} and the target RX latent space, which is denoted by $\y\in\mathbb{C}^{\frac{m}{2}}$ (i.e., the complex version of the RX latent vectors associated with training data without considering channel impairments), with respect to linear semantic precoder and decoder. This facilitates the alignment of the latent spaces while introducing robustness against channel effects. While several distance metrics can be exploited in this formulation, in the sequel, we use mean-squared error as a simple yet effective way to measure latent space mismatches. In formulas, this translates to:
\begin{align}\label{pr:biconvex_min}
\min_{\G,\F}&\quad \mathbb{E}||\y-(\G\H\F\x+\G\v)||^2_2 \nonumber \\
\text{s.t.}&\quad\text{tr}(\F\F^H)\le P_T,
\end{align}
where $\mathbb{E}$ denotes the expected value over the noise and data distributions. Assuming to have $n=|\mathcal{P}|$ i.i.d. training samples $\{(\x^{(i)},\y^{(i)})\}_{i=1}^n$ and leveraging the zero-mean property of the AWGN, the objective function of \eqref{pr:biconvex_min} can be approximated using the empirical loss
\begin{align}\label{pr:biconvex_min_1}
\frac{1}{n}\sum_{i=1}^n||\y^{(i)}-(\G\H\F\x^{(i)})||^2_2 + \text{tr}(\G\Sigma_\v\G^H),
\end{align}
with $\Sigma_\v$ denoting the noise covariance matrix. Let us define $\X\in\mathbb{C}^{\frac{d}{2}\times n}$ be the matrix that collects all training samples, where each column $\x_i$ represents an individual input,
and $\Y\in\mathbb{C}^{\frac{m}{2}\times n}$ be the matrix containing the corresponding receiver latent column vectors. %for each training sample, where each column $\y_i$ represent an individual target output.
The optimization problem \eqref{pr:biconvex_min} can be rewritten in matrix form as: 
\begin{align}\label{pr:biconvex_matrix}
\min_{\G, \F} &\quad \frac{1}{n} || \Y - \G\H\F\X ||_F^2 + \text{tr}\left(\G\Sigma_{\v}\G^H\right) \nonumber \\
\text{s.t.}&\quad\text{tr}(\F\F^H)\le P_T.
\end{align} 
Given the bi-convex nature of problem \eqref{pr:biconvex_matrix}, an efficient solution involves constructing an iterative algorithm that alternates between the solutions of $\G$ and $\F$. To this aim, we reformulate the problem to be amenable for ADMM solution, which allows us to efficiently handle the constraints and solve the bi-convex optimization program in an efficient manner \cite{boyd2011distributed}. Then, let us define $\Omega$ as the set of elements that satisfy the given constraint:
\begin{equation}\label{eq:Omega}
    \Omega=\{\F\ |\ \text{tr}(\F\F^H)\le P_T\}.
\end{equation}
The indicator function $I_\Omega(\F)$ for the set $\Omega$ is defined as:
\begin{equation}\label{eq:indicator}
I_\Omega(\F)=\begin{cases}
+\infty & \text{if } \F\notin \Omega,\\
0 & \text{if } \F\in \Omega.
\end{cases}
\end{equation}
Thus, using \eqref{eq:indicator}, and introducing an auxiliary variable $\Z=\F$, problem \eqref{pr:biconvex_min} can be equivalently rewritten as:
\begin{align}
\min_{\G,\F,\Z}&\quad\frac{1}{n} || \Y - \G\H\F\X ||_F^2 + \text{tr}\left(\G\Sigma_{\v}\G^H\right)  + I_\Omega(\Z) \label{eq:prob_ref}\\ 
\text{s.t.}&\quad \F-\Z=\mathbf{0}. \nonumber
\end{align}
Problem (\ref{eq:prob_ref}) can now be solved using the scaled version of ADMM \cite{boyd2011distributed}, which hinges on the following iterative updates:
\begin{align}
\G^{(t+1)} =&\, \arg\min_\G \;\frac{1}{n} || \Y - \G\H\F^{(t)}\X ||_F^2 \nonumber \\
& + \text{tr}\left(\G\Sigma_{\v}\G^H\right)  \label{eq:g_step}\\ 
\F^{(t+1)} =&\, \arg\min_\F \;\frac{1}{n}||\Y-\G^{(t+1)}\H\F\X||_F^2 \nonumber\\  
&+\rho||\F-\Z^{(t)}+\U^{(t)}||^2_F \label{eq:f_step}\\
\Z^{(t+1)} =&\; \text{Proj}_\Omega(\F^{(t+1)}+\U^{(t)})\label{eq:z_step}\\ 
\U^{(t+1)} =&\; \U^{(t)} + \F^{(t+1)} - \Z^{(t+1)}\label{eq:u_step}
\end{align}
where $t$ denotes the iteration index, $\U$ is the Lagrange multiplier used to enforce the (matrix) constraint in (\ref{eq:prob_ref}), $\rho$ is a positive regularization parameter, and $\text{Proj}_\Omega$ denotes the orthogonal projection onto the set $\Omega$ in \ref{eq:Omega}. The optimization procedure is initialized with $\F^{(0)} \sim \mathcal{CN}(0,1)$, and both $\Z^{(0)}$ and $\U^{(0)}$ are set to zero matrices. In the sequel, we give closed-form expressions for the ADMM steps in \eqref{eq:g_step}-\eqref{eq:z_step}. 

\noindent\textbf{The $\G$-step. } At the $(t+1)$-th iteration, with $\F$ fixed, the update for $\G$ can be obtained by setting the gradient of \eqref{eq:g_step} with respect to $\G^H$ to zero. Recasting the objective of \eqref{eq:g_step} as
\begin{align}\label{eq:simpl_obj}
&\frac{\text{tr}\left((\Y-\G\H\F^{(t)}\X)(\Y-\G\H\F^{(t)}\X)^H\right)}{n}+\text{tr}\left(\G\Sigma_{\v}\G^H\right),
\end{align}
simple algebra leads to the optimal update:
\begin{equation}
\G^{(t+1)}=\Y(\H\F^{(t)}\X)^H\left((\H\F^{(t)}\X)(\H\F^{(t)}\X)^H+n\Sigma_{\v}\right)^{-1}. \label{eq:g_step_final}
\end{equation}
\smallskip
\noindent\textbf{The $\F$-step. }With $\G$, $\Z$ and $\U$ fixed, hinging again on expression \eqref{eq:simpl_obj}, the update for $\F$ at the $(t+1)$-th iteration can be obtained by setting to zero the gradient of the objective function in \eqref{eq:f_step} with respect to $\F^H$ and solving it in closed form. The derivations lead to the following equation:
\begin{align}\label{eq:F_equation}
%\textcolor{red}{\frac{\text{tr}\left((\Y - \G\H\F\X)(\Y - \G\H\F\X)^H\right)}{n}+\rho (\F-\Z+\U).}\\
%(\G\H)^H(\G\H)\F(\X\X^H)+\textcolor{red}{n}\rho \F = \textcolor{red}{n}\rho(\Z-\U)+(\G\H)^H\Y\X^H. 
(\G^{(t+1)}\H)^H(\G^{(t+1)}\H)\F(\X\X^H)+n\rho \F &\nonumber\\ 
\qquad-n\rho(\Z^{(t)}-\U^{(t)})-(\G^{(t+1)}\H)^H\Y\X^H &= 0,
\end{align}
%where $n$ is the total number of training signals.
Now, letting 
\begin{align}
 & \A=(\G^{(t+1)}\H)^H(\G^{(t+1)}\H), \label{eq:A}\\
 & \B=\X\X^H,\label{eq:B}\\
 & \C=n\rho(\Z^{(t)}-\U^{(t)})+(\G^{(t+1)}\H)^H\Y\X^H, \label{eq:C}
\end{align}
we can compactly write \eqref{eq:F_equation} as:
\begin{equation}\label{eq:F_equation2}
    \A\F\B + n\rho\F =\C.
\end{equation}
Then, exploiting
%\begin{equation}\label{eq:kron_intro}
$\text{vec}(\A\F\B)=(\B^H\otimes \A)\text{vec}(\F)$
%\end{equation}
in \eqref{eq:F_equation2} and solving for $\text{vec}(\F)$, we obtain
\begin{equation}
\text{vec}(\F) = (\B^H\otimes \A + n\rho \mathbf{I} )^{-1}\text{vec}(\C),\nonumber
\end{equation}
where $\text{vec}(\cdot)$ denotes the vectorization operator. Finally, the update for $\F$ at the $(t+1)$-th step is:
\begin{equation}
\F^{(t+1)} = \text{vec}^{-1}\left((\B^H\otimes \A + n\rho \mathbf{I} )^{-1}\text{vec}(\C)\right).\label{eq:f_step_final}
\end{equation}
with $\A$ and $\C$ depending on $\G^{(t+1)}$, $\Z^{(t)}$ and $\U^{(t)}$ as in \eqref{eq:A}-\eqref{eq:C}.
To avoid the unfavorable scaling of the Kronecker product for large matrices, \eqref{eq:F_equation2} can be also recasted as a Sylvester equation, which can be efficiently solved using the Bartels-Stewart algorithm \cite{Bartels1972Algorithm4}.%, implemented in Python via the \texttt{scipy} library \cite{2020SciPy-NMeth}.

\smallskip
\noindent \textbf{The $\Z$-step. }Applying the definition of projection operator, the step in \eqref{eq:z_step} can be recast as:
\begin{align}
\Z^{(t+1)} &=\, \text{Proj}_\Omega(\F^{(t+1)}+\U^{(t)})\nonumber\\
%&=\,\arg\min_\Z\,\frac{1}{2}||\F^{(k+1)}+\U^{(k)}-\Z||^2_F + I_\Omega(\Z)\nonumber\\
&=\,\arg\min_\Z \,||\F^{(t+1)}+\U^{(t)}-\Z||^2_F\\ \nonumber
&\qquad\;\quad \text{s.t.}\quad \text{tr}(\Z\Z^H)\le P_T
\end{align}
This problem is convex, and its optimal solution can be found by imposing the Karush-Kuhn-Tucker (KKT) conditions \cite{ghojogh2021kktconditionsfirstordersecondorder}:
\begin{align}
\nabla_{\Z^H}||\F^{(t+1)}+\U^{(t)}-\Z||^2_F + \lambda(\text{tr}(\Z\Z^H)-P_T)=\mathbf{0}\label{eq:Grad_Lagrangian}\\ \nonumber
\text{tr}(\Z\Z^H)-P_T \le 0\\ \nonumber
\lambda \ge 0\\ \nonumber
\lambda(\text{tr}(\Z\Z^H)-P_T)=0 \nonumber
\end{align}
% The Lagrangian is given by:
% \begin{equation}
% \L(\Z,\lambda)=||\F^{(k+1)}+\U^{(k)}-\Z||^2_F + \lambda(\text{tr}(\Z\Z^H)-c)
% \end{equation}
Taking the gradient with respect to the conjugate transpose of the Lagrangian function in (\ref{eq:Grad_Lagrangian}), we obtain:
\begin{align}
% -(\F^{(k+1)}+\U^{(k)}-\Z) + \lambda \Z = 0\\ \nonumber
% (1+\lambda)\Z=\F^{(k+1)}+\U^{(k)}\\ \nonumber
\Z=\frac{1}{1+\lambda}(\F^{(t+1)}+\U^{(t)}).
\end{align}
Now, letting $\hat{\Z}=\F^{(t+1)}+\U^{(t)}$, and recalling that $$\text{tr}(\Z\Z^H)=\frac{1}{(1+\lambda)^2}\text{tr}(\hat{\Z}\hat{\Z}^H),$$ the optimal $\lambda$ is given by:
\begin{align}
\hat\lambda = \begin{cases}
    0, &\quad \text{if tr}(\hat{\Z}\hat{\Z}^H)\le P_T,\\
    \sqrt{\displaystyle\frac{\text{tr}(\hat{\Z}\hat{\Z}^H)}{P_T}}-1, &\quad \text{else.}
\end{cases}
\end{align}
Thus, the update for $\Z$ at the $(t+1)$-th step is:
\begin{equation}
   \Z^{(t+1)} = \frac{1}{1+\hat\lambda}(\F^{(t+1)}+\U^{(t)}).\label{eq:z_step_final}
\end{equation}

\smallskip
\noindent \textbf{The $\U$-step. }Finally, the update for $\U$ at the $(t+1)$-th step is given by the recursive expression in \eqref{eq:u_step}.
% \begin{equation}
%     \U^{(k+1)} = \U^{(k)} + \F^{(k+1)} - \Z^{(k+1)}.
% \end{equation}

The linear model in \eqref{eq:linear_MIMO_x} offers simplicity, low complexity, and a robust training procedure based on ADMM. However, general nonlinear models can typically outperform linear ones, with a price paid in terms of complexity and interpretability. In the next paragraph, we introduce a nonlinear MIMO semantic precoding and decoding strategy based on DNNs.

\begin{algorithm}[t]
\caption{ADMM}
\label{alg:admm_algo}
\begin{algorithmic}[1]
\STATE \textbf{Input:} $ \F^{(0)} $$\sim$$ \mathcal{CN}(0,1) $, $\Z^{(0)} = \U^{(0)} = \mathbf{0}$,  $ n $,  $\Sigma_{\v}$, and $\rho$.
\STATE \textbf{Output:} Final values \( \F^{(T)}, \G^{(T)}, \Z^{(T)}, \U^{(T)} \).
\FOR{each iteration \( t = 1, \dots, T \)}
    \STATE Update $\G^{(t+1)}$ as in \eqref{eq:g_step_final}.
    \STATE Update $\F^{(t+1)}$ as in \eqref{eq:f_step_final}.
    \STATE Update $\Z^{(t+1)}$ as in \eqref{eq:z_step_final}.
    \STATE Update $\U^{(t+1)}$ as in \eqref{eq:u_step}.
\ENDFOR
\end{algorithmic}
\end{algorithm}

\subsection{Neural Semantic Precoding and Decoding}
\label{sec:nn_formulation}

In this section, we introduce the neural network-based approach. Unlike the linear case, the semantic encoding and decoding functions are modeled as nonlinear parametric functions, such as DNNs. Specifically, we have a semantic neural precoder $f_{\boldsymbol{\theta}}(\cdot)$, and a semantic neural decoder $g_{\boldsymbol{\psi}}(\cdot)$, with a set of learnable and complex parameters $\boldsymbol{\theta}$ and $\boldsymbol{\psi}$, respectively. In this setting, we have a general observation model as in \eqref{eq:problem_setting}. The structure of the DNNs can be chosen according to different design choices, and the achievable performance depends on the specific latent spaces to be aligned, and the complexity constraints on the overall architecture. Furthermore, due to the specific complex nature of the signal, we adopt activation functions that operate on complex vectors. In particular, we exploit the phase-amplitude method proposed in \cite{trabelsi2018deepcomplexnetworks}, where a generic activation function $\alpha(\cdot)$ is applied to the magnitude of the complex number, while the phase remains unchanged. Specifically, for a general complex input $\mathbf{z}\in\mathbb{C}^k$, the activation function is defined as
$\varphi(\mathbf{z})=\alpha(|\mathbf{z}|)\cdot e^{j\phi}$,
where $\phi=\tan^{-1}(\frac{\mathbf{b}}{\mathbf{a}})$ represents the phase of the complex vector $\mathbf{z}=\mathbf{a}+j\mathbf{b}$, with $\mathbf{a}, \mathbf{b} \in \mathbb{R}^k$. Finally, the parameters of the neural semantic precoders and decoders are optimized according to the following training criterion:
\begin{align}\label{pr:neural_prob}
\min_{\boldsymbol{\theta},\boldsymbol{\psi}}\;\mathbb{E}||\y-(g_\psi(\H f_\theta(\x)+\v))||^2 + \beta||\boldsymbol{\theta}||_0+\gamma||\boldsymbol{\psi}||_0.
\end{align}
with $\beta,\gamma\geq 0$, where the regularization terms $\beta||\boldsymbol{\theta}||_0$ and $\gamma||\boldsymbol{\psi}||_0$ are used to promote sparsity of the model parameters at the TX and RX side, which reduces memory requirements for model storage and computational complexity at inference time. The presence of noise $\v$ is addressed directly during the forward pass by adding random samples drawn from $\mathcal{CN}(\mathbf{0}, \Sigma_\v)$ to the receiver signal $\H f_\theta(\x)$. This makes the model naturally robust to the noise introduced by the communication channel. The transmitting power constraint is incorporated directly into the neural network structure by implementing $ \ell_2 $ normalization of the semantic encoder output. %Additionally, we promote sparsity in the model parameters by incorporating the term $\gamma||\theta||_0$ to the objective function \cite{louizos2017learning}, where $\gamma$ serves as the regularization parameter. This help reducing the spatial complexity required for storing the neural network model in memory and to reduce the computational complexity at inference time.

We solve problem \eqref{pr:neural_prob} using a Proximal Gradient Descent (PGD) \cite{parikh2014proximal} method with Hard Thresholding, which is described in Algorithm \ref{alg:pgd_alg}. Specifically, after each training epoch, a hard-thresholding operator $ \mathcal{H}_\tau(\cdot) $ is applied to the stacked weight vector $ \w := [\boldsymbol{\theta}, \boldsymbol{\psi}]$ in the following manner:
\begin{align}
\mathcal{H}_\tau(w_{i}) = 
\begin{cases} 
w_{i}, & \text{if } | w_{i} | > \tau_i, \\
0, & \text{otherwise}.
\end{cases}    
\end{align}
where $ \{\tau_i\}_{i=1}^{|\w|} $ is a set of thresholds where each element is defined as:
\begin{align}\label{eq:thresh}
    \tau_i = \begin{cases}
    \beta \cdot \eta, & \quad\text{if }\w_i \in \boldsymbol{\theta};\\
    \gamma \cdot \eta, & \quad\text{if }\w_i \in \boldsymbol{\psi}.
    \end{cases}
\end{align}
with $ \eta $ being the learning rate.

\begin{algorithm}[t]
\caption{PGD with Hard Thresholding}
\label{alg:pgd_alg}
\begin{algorithmic}[1]
\STATE \textbf{Input:} % Initial weights $\w^{(0)}$ and the set of thresholds $\{\tau_i\}_{i=1}^{|\w|}$.
Initial weights $\w^{(0)}$ and $\{\tau_i\}_{i=1}^{|\w|}$.
\STATE \textbf{Output:} The final sparse weights $\w^{(T)}$.
\FOR {each epoch $t\in\{1,\dots,T\}$}
    \STATE Gradient descent update and global hard thresholding:
    \begin{align*}
        \w^{(t+1)} = \mathcal{H}_\tau(\w^{(t)} - \eta \nabla \mathcal{L}(\w^{(t)}));
    \end{align*}
    \vspace{-0.5cm}
    \STATE Freeze gradients of pruned weights:
    \vspace{-0.1cm}
    \begin{align*}
        \nabla \mathcal{L}(w_{i}) = 0, \quad \text{if } w^{(t+1)}_{i} = 0;
    \end{align*}
    \vspace{-0.7cm}
\ENDFOR
\end{algorithmic}
\end{algorithm}

\vspace{-0.5cm}

\subsection{Computational Complexity Analysis}

%Neural networks typically deliver better performance than linear models, but at the cost of increased overall complexity. These models are often overparameterized, making them unsuitable for devices with limited computational resources. In contrast, linear models offer a less complex alternative, although they generally cannot achieve the same level of performance as their neural counterparts. 

In this section, we measure the complexity of the models in terms of the number of floating-point operations (FLOPs) required for computations. We consider each arithmetic operation to count as one FLOP.

Precisely, to compute the multiplication of a matrix $\A\in\mathbb{R}^{a\times b}$ and a vector $\b\in\mathbb{R}^b$, $a(2b-1)$ FLOPs are required. When we consider the bias component, which involves adding a vector $\c\in\mathbb{R}^a$, the FLOPs increase by $a$, resulting in a total $2ab$ FLOPs. When $\A$, $\b$, and $\c$ are complex, the required FLOPs are $a(8b-2)$ for the matrix-vector multiplication and $8ab$ considering the bias too. %This is because a single complex number multiplication requires 4 real multiplications and 2 real additions, while a complex number addition requires 2 real additions. For each of the $b$ elements in the vector, this translates to $6b$ real operations for multiplication and $2(b-1)$ real operations for addition, resulting in $8b-2$ FLOPs for each of the $a$ rows.
%Now, we will consider sparse matrices to reduce the number of operations and increase storage efficiency. In particular, we will consider only non-zero elements contribute to the FLOP count.
%Thus, when the matrix $ \A $ becomes sparse, the number of required FLOPs reduces to $ 8 \cdot \text{nnz}(\A) - 2 \cdot \text{nnzrow}(\A) $, where $ \text{nnz}(\A) $ represents the total number of non-zero elements in $ \A $, and $ \text{nnz\_row}(\A) $ denotes the number of rows in $ \A $ containing at least one non-zero element.
%For a sparsity level $ s \in [0, 1] $, the total number of non-zero elements is given by $ \text{nnz}(\A) = (1 - s)ab $. Determining the number of non-zero rows, $ \text{nnzrow}(\A) $, is not so trivial. However, under the assumption of uniform sparsity, where zeros are distributed uniformly at random, it can be observed that $ \text{nnzrow}(\A) \geq \text{rank}(\A) $. Additionally, the rank of $ \A $ remains $ \text{rank}(\A) = a $, even for high sparsity levels such as $ s \approx 0.9 $. This allows us to infer that $ \text{nnzrow}(\A) = a $ for high sparsity levels. Therefore, the total number of FLOPs required for sparse $ \A $ can be expressed as $a(8(1 - s)b - 2)$.
We explicitly incorporate matrix sparsity into our DNN models to reduce computational requirements and improve storage efficiency. The sparsity level is defined as $s \in [0, 1]$, where $s$ represents the fraction of zero elements in the matrix $\A$. Thus, the total number of non-zero elements in $\A$ is assumed to be $(1-s)ab$. Consequently, the number of FLOPs required for a forward pass through a complex linear layer can be approximated as $8(1-s)ab$. This formulation demonstrates how sparsity effectively reduces computational costs. For the sake of clarity, a summary of FLOPs calculations is provided in Table \ref{tab:FLOPs_summary}.
\begin{table}[t]
    \centering
    \caption{Approximated FLOP counts for the considered cases.}
    \label{tab:FLOPs_summary}
    \begin{tabular}{lllll}
    \hline
    \textbf{Case}               & \textbf{Complex} & \textbf{Sparse} & \textbf{Dense} & \textbf{FLOPs}        \\ \hline
    $\A\b$        & \checkmark       &                  & \checkmark      & $a(8b-2)$              \\
    $\A\b+\c$     & \checkmark       &                  & \checkmark      & $8ab$                  \\
    $\A\b+\c$    & \checkmark       & \checkmark       &                  & $8(1 - s)ab$           \\ \hline
    \end{tabular}
\end{table}

\noindent\textbf{Linear model FLOPs.} Consisting of the implementation of two complex dense matrices $ \F \in \mathbb{C}^{K N_T \times \frac{d}{2}} $ and $ \G \in \mathbb{C}^{\frac{m}{2} \times K N_R} $, the total number of FLOPs required for the linear model is:
\begin{align}\label{pr:linear_FLOPs}
\#_\text{FLOPs} &= K N_T\left(8\frac{d}{2}-2\right) + \frac{m}{2}\left(8KN_R-2\right) \nonumber \\
&= 4KN_Td+(4KN_R-1)m-2KN_T
\end{align}

\noindent\textbf{Neural model FLOPs.} %The total FLOPs for the neural model includes contributions from the input layer, hidden layers, and output layer for both the semantic precoder and decoder. 
Let $ i_p $, $ o_p $, $ h_p $, and $ l_p $  be the number of neurons in the input layer, output layer, hidden layers, and the number of hidden layers for the precoder, respectively. Similarly, we have $ i_d $, $ o_d $, $ h_d $, and $ l_d $ for the decoder side. Then, given a level of sparsity $ s $, the total number of FLOPs required by the neural network solution, under the assumption of uniform sparsity in the weights, is given by:
\begin{align}\label{pr:neural_sparse_FLOPs}
\#_\text{FLOPs} =&\ 8(1-s)[h_pi_p+o_ph_p+h_di_d+o_dh_d+(l_p-1)h_p^2\nonumber\\
&+(l_d-1)h_d^2 ] + c(l_ph_p+l_dh_d)
\end{align}
%Then, the total number of FLOPs required by the neural model is:
%\begin{align}\label{pr:neural_dense_FLOPs}
%\#_\text{FLOPs} =&\ 8[h_pi_p+o_ph_p+h_di_d+o_dh_d\nonumber\\
%&\hspace{-.4cm}+(l_p-1)h_p^2+(l_d-1)h_d^2 ]+ k(l_ph_p+l_dh_d)
%\end{align}
where $ c $ represents the FLOPs required by the complex activation function $ \varphi(\cdot) $.

% \smallskip
% \noindent\textbf{Sparse neural model FLOPs.} Given a level of sparsity $ s $, the total number of FLOPs required by the neural network solution, under the assumption of uniform sparsity in the weights, is given by:
% \begin{align}\label{pr:neural_sparse_FLOPs}
% \#_\text{FLOPs} =&\ 8(1-s)[h_pi_p+o_ph_p+h_di_d+o_dh_d\nonumber\\
% &+(l_p-1)h_p^2+(l_d-1)h_d^2 ]\nonumber\\
% &+ k(l_ph_p+l_dh_d)
% \end{align}

\section{Numerical Results}

\begin{figure}[t]
    \centering    \includegraphics[width=\columnwidth, trim=5bp 5bp 5bp 5bp, clip]{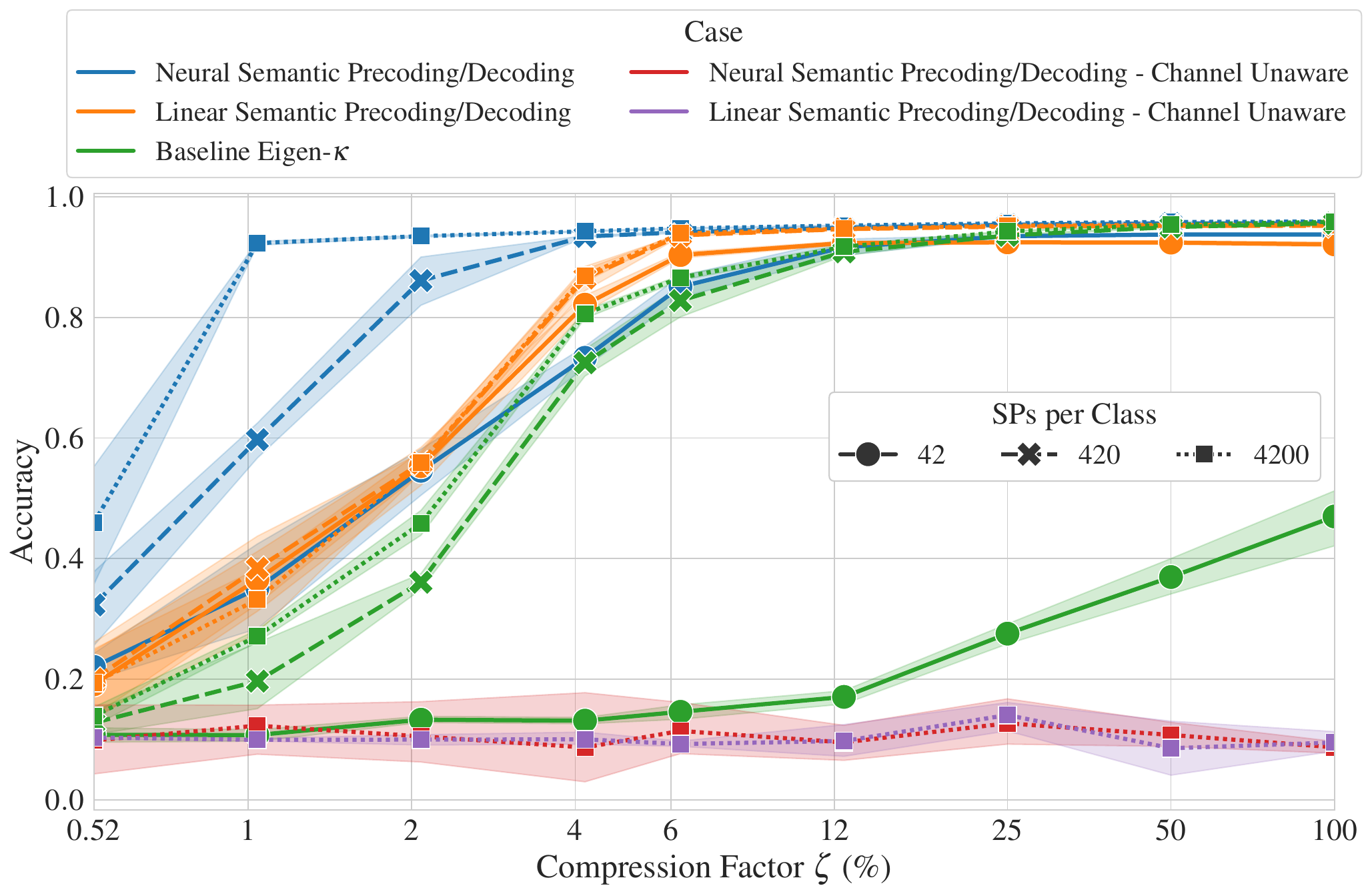}
    \caption{Accuracy versus $\zeta$, with ${\rm SNR}=20$ dB and $\beta=\gamma=0$.}
    \label{fig::accuracy}
    %\vspace{-0.6cm}
\end{figure}

\begin{figure}[t!]
    \centering    \includegraphics[width=\columnwidth, trim=5bp 5bp 5bp 5bp, clip]{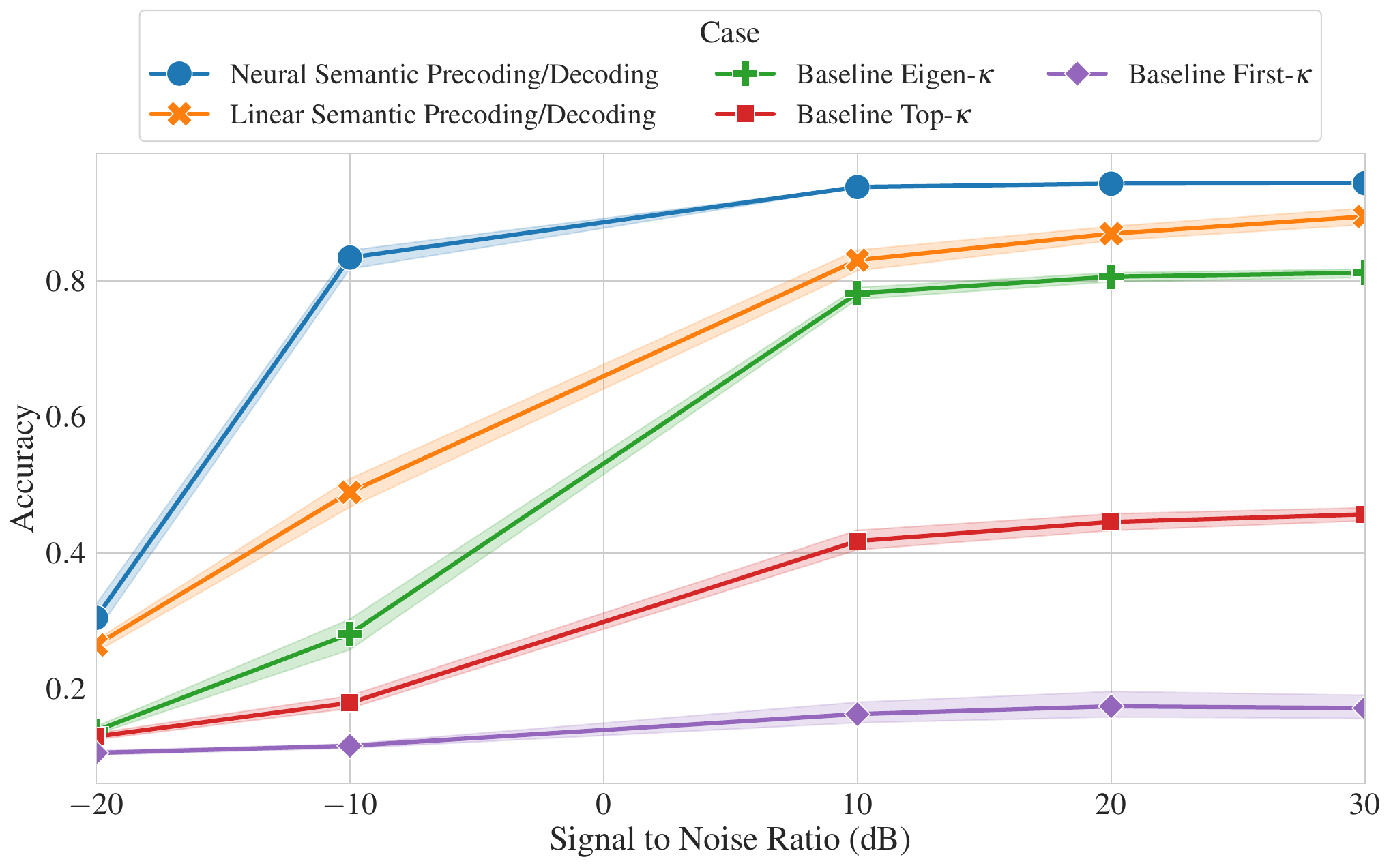}
    \caption{Accuracy versus SNR, with $\zeta\approx 4\%$ and 4,200 SPs per class.}
    \label{fig::snr}
\end{figure}

\begin{figure}[t!]
    \centering    \includegraphics[width=\columnwidth, trim=5bp 5bp 5bp 5bp, clip]{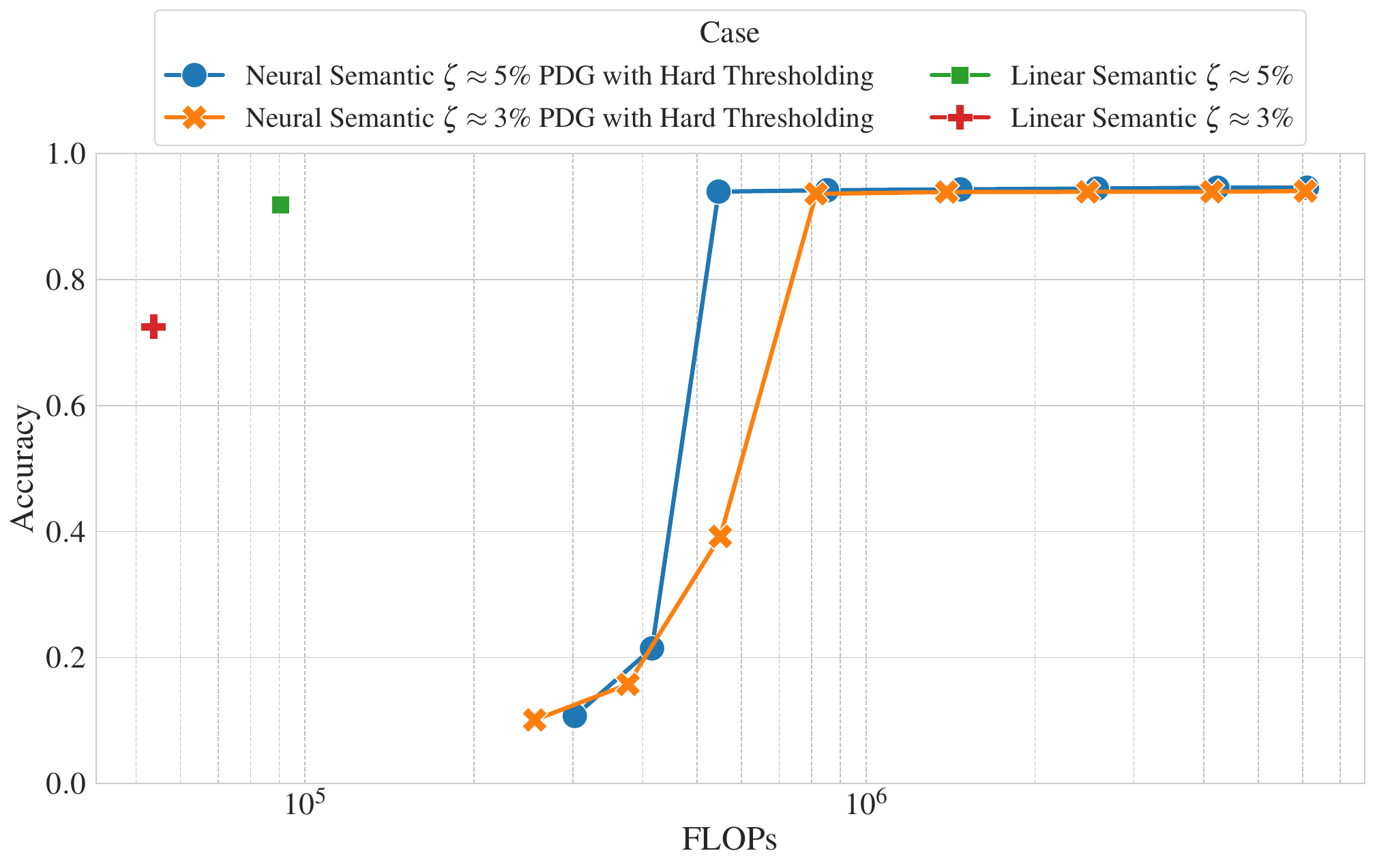}
    \caption{Accuracy versus FLOPs, with $ \zeta \approx 3\% $ and $ \zeta \approx 5\% $, ${\rm SNR} = 20$ dB and 4,200 SPs per class.}
    \label{fig::FLOPs}
    %\vspace{-0.4cm}
\end{figure}

In this section, we evaluate the performance of the proposed semantic equalization methods through numerical experiments. We used the CIFAR-10 dataset \cite{krizhevsky2009learning}, comprising 60,000 $32$$\times$$32$ color images distributed across 10 classes. % (6,000 images per class). 
%Among them, 42,500 images were used for training, from which subsets of 420, 4,200, and 42,000 images were sampled uniformly at random as SPs, 7,500 for validation, and 10,000 for testing, with classification across 10 labels as the downstream task, $\mathcal{T}$. 
Among them, 42,500 images were used for training, from which we uniformly sample three subsets with balanced class distributions, resulting in a total of $n\in\{420, 4200, 42000\}$ SPs. The remaining 7500 images are used for validation, and 10000 for testing, with classification across 10 labels as the downstream task, $\mathcal{T}$. 
Furthermore, we employed two encoding models from the Python library \textit{timm} \cite{rw2019timm}: (i) The \textit{vit\_small\_patch16\_224} model to generate the transmitter-side encodings with $d = 384$; (ii) the \textit{vit\_base\_patch16\_224} model for the receiver-side encodings with $m = 768$. For simplicity, we consider square MIMO Rayleigh fading channels with unitary variance, where the number of transmitting and receiving antennas are equal, i.e., $N_T = N_R$. All results were averaged across six random seeds%(27, 42, 100, 123, 144, 200)
, considering  $\rho=100$. \footnote{\href{https://github.com/SPAICOM/semantic-alignment-mimo.git}{https://github.com/SPAICOM/semantic-alignment-mimo.git}}\\
The neural semantic precoder and decoder are implemented using a multilayer perceptron architecture, each with a single hidden layer ($ l_p = l_d = 1 $), consisting of $ h_p = \frac{d}{2} = 192 $ and $ h_d = \frac{m}{2} = 384 $ neurons, respectively. % The total number of neurons in the semantic precoder is $ 2 \cdot \frac{d}{2} + K N_T = 384 + K N_T $, while for the semantic decoder it is $ 2 \cdot \frac{m}{2} + K N_R = 768 + K N_R $.
% The FLOPs complexity for the dense implementation is:
% \begin{align}\label{pr:neural_dense_FLOPs_final}
% \#_\text{FLOPs} =&\ 2(d^2 + m^2) + 2(2N_T + k)d \nonumber \\
% &+ 2(2N_R + k)m,
% \end{align}
% where $ k $, representing the FLOPs required by the activation function, is set to $ 100 $. For the sparse implementation, 
From \eqref{pr:neural_sparse_FLOPs}, the FLOPs complexity reduces to $
\#_\text{FLOPs} =\ 2(1-s)\big[d^2 + m^2 + 2K (N_T d + N_R m)\big]+ \frac{c}{2}(d + m)$. The linear model was trained for 20 iterations, while the neural network was trained for 50 epochs with a learning rate of $ \eta = 10^{-3} $.
In the neural model, we set $\beta$$=$$\gamma$ to set a global threshold $\tau$.
%the coefficients $\beta$ and $\gamma$ are set to be equal, allowing the unification of the thresholds $\{\tau_i\}_{i=1}^{|\w|}$ in \eqref{eq:thresh} into a single global threshold $\tau$. 
\\
\indent \textbf{Baselines.} We compare our methodology with baselines that directly transmit vectors $\x$ of the TX latent space while performing semantic alignment and MIMO channel equalization in a disjoint fashion. Specifically, letting $\overline{\H} = \U\S\V^H$, the baselines involve SVD-based MIMO channel equalization, where precoding and decoding are chosen as:
\begin{align}
\F &= \mathbf{1}_K\otimes\V,\\
\G &= \mathbf{1}^T_K\otimes\left(\S^H\S+\frac{1}{\text{SNR}} \mathbf{I}_{N_R}\right)^{-1}(\U\S)^H,
\end{align}
where $\mathbf{1}_K$ is the $K$-dimensional unitary vector, and $\text{SNR}$ represents the Signal-to-Noise Ratio (SNR) of the communication channel.
%Once the TX latent vectors $\x$ are received, semantic alignment is performed to map them into the RX latent space using a linear transformation obtained by solving the following least-square problem:
%\textcolor{red}{Semantic alignment is performed by mapping the TX latent space into the RX one via a linear transformation $\A\in\mathbb{R}^{m\times d}$ such that $\mathbf{s}_R = \A^{\dagger}\mathbf{s}_T$ obtained by solving a least-square problem.}
Semantic alignment is performed by mapping the TX latent space into the RX one via a linear transformation obtained by solving the following least-square problem:
\begin{align}\label{pr:least_square}
\min_{\Q\in\mathbb{R}^{m\times d}}\frac{1}{n}\sum_{i=1}^{n}||{\mathbf{s}_R}^{(i)}-\Q\mathbf{s}_T^{(i)}||_2^2.
\end{align}
To assess the performance of the baselines for different amounts of transmitted information, we consider three alternative strategies: (i) the First-$\kappa$ baseline (First-$\kappa$), which transmits the first $2K N_T$ features of $ \mathbf{s}_T$; (ii) the Top-$\kappa$ baseline (Top-$\kappa$), which selects and transmits the largest $K N_T$ features of $\mathbf{s}_T$ in terms of $\ell_1$-norm; and (iii) the Eigen-$\kappa$ baseline (Eigen-$\kappa$) in which the TX latent space is first precoded using $\tilde{F}=\tilde{\Sigma}_QV_Q^T$ and then after MIMO equalization via SVD the RX latent space is decoded using $\tilde{G}=U_Q\tilde{\Sigma}_Q$, where $U_Q\Sigma_Q V_Q^T$ is the singular value decomposition of the matrix $\Q$ in \eqref{pr:least_square}, and $\tilde{\Sigma}_Q$ is a filtered version of the $\Sigma_Q$ in which only the largest $2 K N_T$ eigenvalues are retained.
For First-$\kappa$ and Top-$\kappa$ baselines, the alignment is performed once the TX latent vectors are received by RX.
Specifically, for Top-$\kappa$ baseline, to ensure accurate signal reconstruction at the receiver, the transmitted signal is structured such that half of the transmission space is allocated to the indices corresponding to the transmitted features, while the remaining half is dedicated to the features themselves. This allocation effectively doubles the communication burden. It is assumed that the indices are perfectly reconstructed at the receiver.
Given $d$, and choosing $K$ and $N_T$, we can set a specific compression ratio in \eqref{eq:compression_factor} for the three baselines.
\\
\indent \textbf{Comparisons.} We illustrate the performance of our approach in Fig. \ref{fig::accuracy}, which shows the average accuracy of the classification task as a function of the compression factor $\zeta$, for varying numbers of SPs. %, sampled uniformly at random.} %for different competing strategies. 
Specifically, we compare: (i) the proposed neural semantic precoding/decoding strategy in Sec. III.B; (ii) The linear semantic precoding/decoding method introduced in Sec. III.A; and (iii) the Eigen-$\kappa$ baseline. %(iii) The First-$\kappa$ baseline; and (iv) The Top-$\kappa$ baseline. 
For the proposed neural and linear methods, we also examine their optimization in the absence of wireless channel knowledge, referring to these strategies as ``channel-unaware.'' The results are obtained considering an SNR equal to 20 dB, and setting $\beta=\gamma=0$. We explore different compression factors $\zeta$ by setting $K=1$ and varying the number of antennas. As we can notice from Fig. \ref{fig::accuracy}, the proposed semantic equalization strategies outperform the Eigen-$\kappa$ baseline, showing excellent performance even at large compression levels. For instance, the proposed neural approach achieves nearly $90\%$ accuracy while transmitting only two complex symbols through the channel, i.e., $\zeta\approx 1\%$ of the original dimension in terms of transmitted symbols. Moreover, at small compression factors, the linear method outperforms the Eigen-$\kappa$ baseline by roughly ten percentage points, demonstrating the advantages of joint over disjoint design. Comparable performance is observed after $\zeta \approx 12\%$, whereas the linear model achieves peak performance at $\zeta \approx 6\%$. %The linear method also achieves the best performance while transmitting a small amount of compressed symbols, i.e., $\zeta$$\approx$$ 6\%$. 
Unsurprisingly, the proposed neural method is more affected by the number of SPs used for alignment, while the linear counterpart remains relatively robust. In contrast, the baseline performs poorly when only a limited number of SPs are available. As expected, the channel-unaware methods perform poorly, further reinforcing the importance of incorporating physical channel information for effective semantic channel equalization in wireless communications.
\\
\indent As a further result, in Fig. \ref{fig::snr}, we illustrate the behaviour of the average accuracy versus the SNR, comparing the proposed strategies with all the baselines. In this simulation, we consider $K$$=$$1$, 4,200 SPs per class and $\zeta \approx 4\%$.  As expected, the accuracy increases at larger SNR, with the neural approach consistently outperforming the linear counterpart. The performance gap previously observed between the linear method and the Eigen-$\kappa$ baseline remains evident across different SNR levels, as illustrated in Fig. \ref{fig::snr}. This consistency suggests that joint optimization not only improves performance under high compression but also enhances robustness to noise, likely by leveraging the channel structure during semantic compression.
Notably, the neural model delivers excellent performance even considering a limited number of antennas, only one channel usage, and low SNR levels. However, this performance improvement comes with a price in terms of computational complexity.  As an example, the neural model requires approximately 113 times more FLOPs than the linear model for $\zeta\approx 3\%$. Nevertheless, NNs are typically overparametrized for the task at hand, and it becomes interesting to investigate the performance-complexity trade-off achievable through sparsification of the NN weights. To this aim, we train the neural model as in (\ref{pr:neural_prob}) using Algorithm 1, varying the hyperparameters $ \beta=\gamma $ across the set $ \{0, 10, 20, 30, 40, 50, 60, 70\} $ to enforce different level of weight sparsity and, consequently, different computational complexity (cf. \ref{pr:neural_sparse_FLOPs}). The results are reported in Fig. \ref{fig::FLOPs}, which illustrates the behavior of the average accuracy versus the number of FLOPs, comparing the linear and neural approaches, for different compression factors, with $\text{SNR}=20$ dB and 4,200 SPs per class. %The sparsified neural model maintains the same accuracy as the original model while requiring only approximately 10\% of the original FLOPs \textcolor{red}{for $\zeta\approx 5\%$}. 
However, as the number of FLOPs approaches that of the linear model, performance degrades significantly, making the linear method optimal under tight budgets, while the neural model is preferable when resources permit.
%. This observation justifies the use of the linear approach under strict computational constraints, whereas the neural model becomes preferable when these constraints are more relaxed, and performance must be prioritized.
\section{Conclusions}
%\vspace{-0.1cm}
This paper introduces a novel framework for latent space alignment in MIMO semantic communications, addressing the challenges posed by both semantic mismatch and channel distortions. Our approach integrates semantic compression with the alignment of semantic spaces through a joint optimization of MIMO semantic precoding and decoding, ensuring robust and efficient transmission. To solve this problem, we propose and evaluate two distinct methodologies. The first is a linear optimization approach, formulated as a biconvex optimization problem and solved using ADMM. The second is a neural network-based solution that learns an optimized mapping between semantic representations and the transmitted signals. 
Numerical results validate the proposed methodologies. % Numerical results demonstrate that the proposed approaches outperform the baselines available from the literature. 
Furthermore, the neural model consistently outperforms the linear one across various channel conditions and noise levels, achieving superior alignment and robustness. However, this performance gain comes at the cost of increased computational complexity. To mitigate this issue, we applied sparsification to the neural network model using Proximal Gradient Descent with Hard Thresholding, significantly reducing the FLOPs while preserving competitive accuracy. However, as the number of FLOPs approaches that of the linear model, performance degrades significantly. This finding highlights the suitability of the linear approach in scenarios with stringent computational limitations, while the neural model emerges as the better choice when computational resources allow for greater flexibility and performance is the primary concern. Future work will focus on further optimizations, leveraging the dynamic and reconfigurable nature of the channel matrix to improve adaptability, or considering more challenging multi-user semantic communication scenarios.

\balance
\bibliographystyle{IEEEbib}
\bibliography{refs}

\end{document}